\documentclass{ifacconf}

\usepackage{graphicx}      % include this line if your document contains figures
\usepackage{natbib}       % required for bibliography

\usepackage{epsfig} % for postscript graphics files
\usepackage{amsmath} % assumes amsmath package installed
\usepackage{amssymb,bm}  % assumes amsmath package installed
\usepackage{mathtools, nccmath}
\usepackage{commath}
\usepackage{subcaption}

\usepackage{algorithm}
\usepackage[noend]{algpseudocode}

\algrenewcommand\algorithmicforall{\textbf{foreach}}
\algrenewcommand\algorithmicindent{.8em}

\usepackage{xcolor}

\begin{document}
\emergencystretch 3em
\begin{frontmatter}

\title{Safety Assurance for Quadrotor Kinodynamic Motion Planning\thanksref{footnoteinfo}}
% \title{Kinodynamic Planning  with Safety Assurance: An Application to Autonomous Quadrotors\thanksref{footnoteinfo}} 

\thanks[footnoteinfo]{This work was supported by University of Houston through a grant.}

\author[First]{Theodoros Tavoulareas}
\author[First]{Marzia Cescon} 
% \author[Third]{Third C. Author}

\address[First]{Department of Mechanical and Aerospace Engineering, University of Houston, Houston TX, 77004 USA (e-mails: ttavoulareas@uh.edu, mcescon2@uh.edu).}
% \address[Second]{Colorado State University, 
%    Fort Collins, CO 80523 USA (e-mail: author@lamar. colostate.edu)}
% \address[Third]{Electrical Engineering Department, 
%    Seoul National University, Seoul, Korea, (e-mail: author@snu.ac.kr)}

\begin{abstract}                % Abstract of 50--100 words
Autonomous drones have gained considerable attention for applications in real-world scenarios, such as search and rescue, inspection, and delivery. As their use becomes ever more pervasive in civilian applications, failure to ensure safe operation can lead to physical damage to the system, environmental pollution, and even loss of human life. Recent work has demonstrated that motion planning techniques effectively generate a collision-free trajectory during navigation. However, these methods, while creating the motion plans, do not inherently consider the \textit{safe operational region} of the system, leading to potential safety constraints violation during deployment. In this paper, we propose a method that leverages run time safety assurance in a kinodynamic motion planning scheme to satisfy the system's operational constraints. First, we use a sampling-based \textit{geometric} planner to determine a high-level collision-free path within a user-defined space. Second, we design a low-level \textit{safety assurance filter} to provide safety guarantees to the control input of a Linear Quadratic Regulator (LQR) designed with the purpose of trajectory tracking. We demonstrate our proposed approach in a restricted 3D simulation environment using a model of the Crazyflie 2.0 drone.
\end{abstract}


\end{frontmatter}
%===============================================================================

\section{INTRODUCTION}
Unmanned aerial vehicles (UAVs), commonly known as drones, are designed to operate autonomously or with minimal human intervention and can perform various tasks and missions previously inaccessible to or dangerous for humans. They provide significant solutions to a wide variety of real-world problems, such as search and rescue, surveying and mapping, warehouse monitoring, delivery, and agriculture (\cite{fahlstrom2022introduction}). In particular, as UAVs are being expected to operate autonomously in critical tasks and missions, in poorly understood environments, and often in the vicinity of humans and other intelligent systems, ensuring that they operate safely and reliably in every circumstance becomes a top priority.

\begin{figure}[htbp]
\centerline{\includegraphics[width=\linewidth]{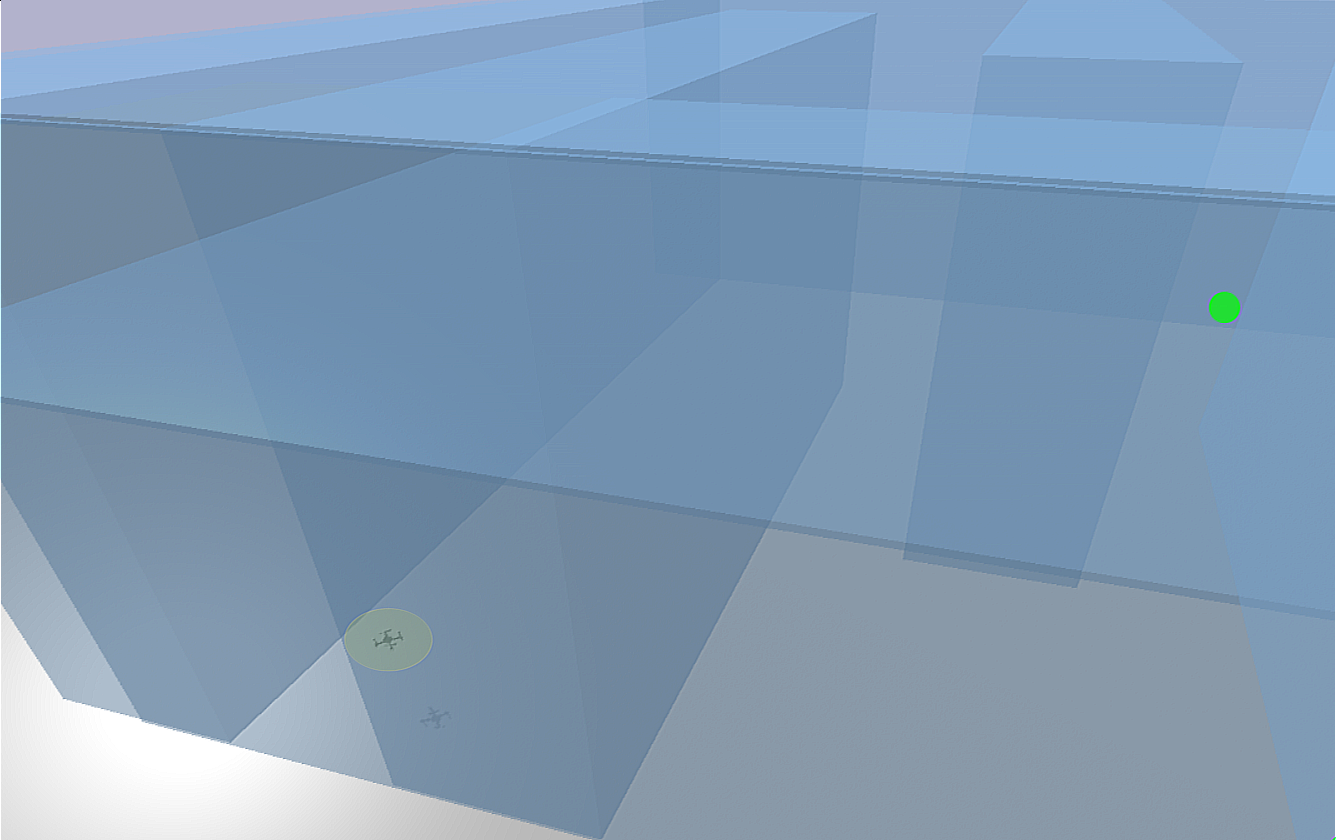}}
% \centerline{\includegraphics[scale=0.25]{env_f.png}}
\caption{The 3D workspace environment with the starting position of the Crazyflie 2.0 quadrotor (yellow sphere), and the target configuration (green sphere). The quadrotor is tasked with reaching the goal by flying between the walls.}
\vspace{0.5em}
\label{fig:env}
\end{figure}

%These tasks often suffer from significant uncertainties in the system dynamics making classical guidance, navigation and control methods less effective. 
The need for safety is, therefore, essential as the research on UAVs is addressing real engineering challenges (\cite{hobbs2023}). In particular, in the context of motion planning with safety guarantees, the planning framework has the objective to guarantee \textit{safety in terms of state conditions} even when the dynamics are fully known. Online planners developed for deterministic environments are insufficient for managing uncertainty, as they typically depend on re-planning strategies or feedback controllers (\cite{guo2023}). Consequently, it is crucial for motion planners to account for uncertainty in real time and respond effectively while ensuring safety, in order to prevent costly hardware failures and enhance convergence (\cite{brunke2022safe}). 
% Integrating data-driven control methods with machine learning has gained popularity for combining the strengths of both methods. However, these approaches lack the flexibility and adaptability across various systems and tasks that learning-based controllers offer~\cite{brunke2022safe}. 

\begin{figure}[htbp]
\centerline{\includegraphics[width=\linewidth]{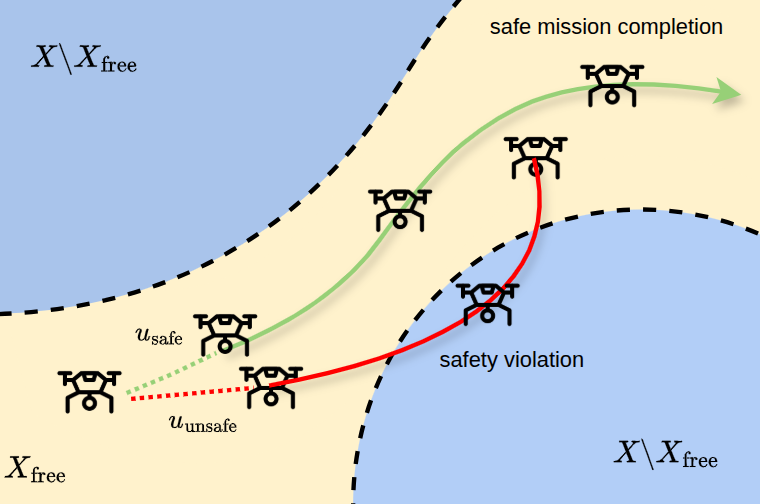}}
\caption{An example where safety problems occur in kinodynamic motion planning. The application of an unsafe control input can result in a safety constraint violation at some point in the future. This is indicated by the red trajectory, where the quadrotor ends up leaving the free 3D space $X_{\text{free}}$ and colliding with the obstacle.}
\vspace{0.5em}
\label{fig:safety}
\end{figure}

In this work, we focus on the problem of providing safety assurance to kinodynamic planning, and in doing so, we focus our attention to quadrotors as a representative system. We consider scenarios such as that shown in Fig.~\ref{fig:env}, in which the quadrotor needs to navigate in a 3D environment and reach the desired goal, without collision and safety violations. Given some safety constraints on the state of the quadrotor determined by the restricted environment, this can be a very challenging task.  For instance, as demonstrated in Fig.~\ref{fig:safety}, a control input applied to the quadrotor to follow a trajectory that is not safe for future states may lead to an aggressive maneuver which jeopardizes the success of the mission. Against this background, in this work, we propose to introduce a safety assurance filter to a geometric motion planner and a primary controller, to detect and adjust control inputs which may lead to safety violation in the future, in order to keep the system safe throughout the mission.
\subsubsection{Contributions}
%Inspired by the mechanism of model predictive control (MPC), we design a safety filter that given a (desired) control input from the LQR that is likely to lead to safe violation in the future, it constructs a safe backup plan towards a safe terminal set in a certain number of steps up to a given finite horizon. At the next time step, the previously computed safe backup plan is available, thus, maintaining safety at all times. In the case where a plan that will move the system further toward the safe terminal set, the safety filter will adjust the desired input of the controller for this to work. The contributions of this work can be summarized in the following way:~\cite{hobbs2023}
We design an approach to safe kinodynamic planning combining a geometric planner with run time assurance provided by an implicit safety filter:
\begin{enumerate}
\item For each time period of a certain length, we propose to generate a geometric collision free path from start to goal with the sampling-based RRT\textsuperscript{*} planner.
\item We exploit a safety assurance mechanism based on predictive control that, given a motion plan from the geometric planner, computes a sequence of future control inputs to be applied to the system with the objective of reaching the goal, that are guaranteed to maintain the system in the region of its safe operation at all future times while reaching a safe terminal set. 
    %\item We propose a straightforward yet effective approach to boost a sampling-based geometric planner (RRT\textsuperscript{*}) along with an optimal control method (LQR) for safety assurance with respect to dynamic constraints.
    %\item To provide safety at all times, a safety filter is presented that checks and verifies the safety of each control input provided by the LQR controller in real-time. This ensures that all control inputs can be safely verified before being applied to the quadrotor system.
\end{enumerate}
The result is an adjusted motion plan compared to that of the geometric planner and primary controller, as the system reaches its goal position, but assures safety in terms of state constraints during navigation. We demonstrate the effectiveness of the proposed framework in a restricted 3D simulation environment using a Crazyflie 2.0 drone model.

%%%%%%%%%%%%%%%%%%%%%%%%%%%%%%%%%%%%%%%%%%%%%%%%%%%%%%%%%%%%%%%%%%%%%%%%%%%%%%%%
\section{RELATED WORK} \label{sec:related}
\subsubsection{Motion Planning}
Kinodynamic planning has historically been a fundamental concept in robotics (\cite{donald1993}). In particular, motion planning for quadrotors is a challenging problem since they are complex systems that are tasked to navigate in highly uncertain environments. For such complex systems, classical feedback control methods can guarantee safety but they may stuck in undesired local minima or struggle with extensive computation time. Also, inaccurate models can lead to suboptimal or unstable behavior. Sampling-based motion planning has been introduced to tackle such scenarios (\cite{orthey2023sampling}). 

The two main categories are graph-based approaches, such as probabilistic roadmaps (PRM) by~\cite{kavraki1996}, for multi-query problems and tree-based, such as rapidly-exploring random trees (RRT) by~\cite{kuffner2000rrt}, expansive space trees (EST) by~\cite{est1997}, and kinodynamic motion planning by interior-exterior cell exploration (KPIECE) by~\cite{csucan2009kinodynamic} for single-query problems. In general, sampling-based planners are easy to design but only probabilistically complete (\cite{choset2005principles}). Moreover, variations of these planners have been introduced that have proven to be asymptotically optimal, but they still suffer from probabilistic completeness. Examples include the RRT\textsuperscript{*} by~\cite{karaman2011sampling}, RRT\textsuperscript{\#} by~\cite{arslan2012role}, and fast-marching trees (FMT\textsuperscript{*}) by~\cite{janson2015fast}. Some of these planners, such as RRT, and EST have been extended to kinodynamic motion planning (\cite{lavalle2001randomized, hsu2002}).

\subsubsection{Safety Assurance}
Run time assurance (RTA) or safety filters can provide safety guarantees to a primary performance-driven controller by intervening and correcting the controller's input for safety when necessary (\cite{miller2024}). Building on this idea, control theoretic frameworks such as control Lyapunov functions (CLFs), evolved into what are now known as Control Barrier Functions (CBFs) (\cite{ames2019}). These frameworks have also been combined with data-driven models for learning tasks~\citep{freire2023}, expanding their application scope. In~\cite{tayal2024}, a cone CBF mechanism is proposed for safe quadrotor navigation in dynamic environments. This constraint formulation utilizes CBFs and collision cones to ensure that quadrotor's relative velocity with respect to obstacles avoids the set of vectors that could result in a collision. In~\cite{tao2023} and~\cite{desa2024}, the authors employ concepts of data-driven CBFs to enhance the robustness of the safety filters during collision avoidance in quadrotor navigation, either using point clouds or predictive control schemes respectively. Although CBF-based safety filters offer a promising approach due to their efficiency and robustness, they are generally difficult to define, they heavily rely on accurate system models and struggle to handle constraints that evolve during operation. 

%Similar to our work, run time assurance (RTA) approaches use a performance-driven primary controller along with an RTA filter that intervenes and corrects the controller's input for safety when necessary~\cite{dunlap2022,miller2024}. 

Additionally, predictive safety filters (PSFs) determine certified control inputs using model predictive control (MPC) (\cite{wabersich2023}). In~\cite{bejarano2023}, a multi-step model predictive safety filter (MPSF) utilizing robust MPC is introduced that significantly reduces chattering in safe control of cartpole and quadrotor systems. This is achieved by incorporating a multi-step horizon in the MPC. In~\cite{chen2024}, an adaptive safety predictive corrector (ASPC) is proposed to provide safe control actions to a reinforcement learning (RL) controller. This method also utilizes a constrained version of MPC.

\begin{figure*}[htbp]
\centerline{\includegraphics[width=\linewidth]{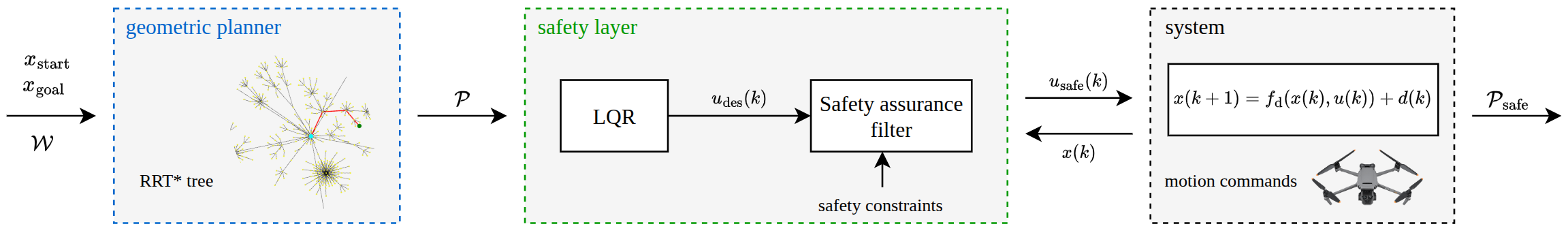}}
\caption{A high-level illustration of the proposed kinodynamic motion planning framework with safety guarantees during one iteration. The geometric RRT\textsuperscript{*} receives as input the environment $\mathcal{W}$, start and goal positions $x_{\text{start}}$ and $x_{\text{goal}}$, respectively, and computes a collision-free path that is, then, turned into a smooth trajectory $\mathcal{P}$ before sending it to the safety layer. In the safety layer, the LQR controller outputs a desired control input $u_{\text{des}}(k)$. The safety filter enforces state constraints and produces a control input $u_{\text{safe}}(k)$ as close as possible to $u_{\text{des}}(k)$ that is applied to the system (where the state $x(k)$ is being fed back into the controller and safety filter) to adjust the planned trajectory to $\mathcal{P}_{\text{safe}}$. We note that this process is repeated for every time horizon $T$, from $x_{\text{start}}$ to $x_{\text{goal}}$, thus, maintaining safety at all times.}
\label{fig:method}
\end{figure*}

That said, in this work, we leverage geometric sampling-based motion planning and optimal control with safety assurance in an end-to-end fashion, thus, advancing toward complete drone autonomy.

% Similarly, \cite{safaoui2024} use chance constraint control methods to ensure safety among multiple drones so they do not collide with each other.

%%%%%%%%%%%%%%%%%%%%%%%%%%%%%%%%%%%%%%%%%%%%%%%%%%%%%%%%%%%%%%%%%%%%%%%%%%%%%%%%
\section{METHODOLOGY} \label{sec:method}
%\subsection{Definitions and Notations}
\subsection{Preliminaries}
% We denote scalars in lowercase $s$, vectors in lowercase bold $\bm{v}$, and matrices in uppercase bold $\bm{M}$. 
We denote the angles and angular velocities of the quadrotor roll ($p,\phi$), pitch ($q,\theta$), and yaw ($r,\psi$), respectively. Also, we denote the workspace with $\mathcal{W}$. In terms of UAV motion planning, the task is to navigate from an initial state to a goal region. The set of all states is called state space, and it is represented as $X$, while its elements are denoted as $x$. The state space is not entirely constraint-free. Some region $X_{\text{free}}$ is free of constraints, while its complement $X \backslash X_{\text{free}}$ represents the total obstacles in the environment (see also Fig.~\ref{fig:safety}).

\subsubsection{Problem Statement (Safe Kinodynamic Planning)}
The problem can be described as a tuple $(X_{\text{free}}, x_{\text{start}}, x_\text{{goal}}, f)$, where $f$ represents the dynamics and kinematics of the system. Therefore, the problem we are concerned with is finding a time-varying trajectory $\mathcal{P}$ from a start state $x_{\text{start}} \in X_{\text{free}}$ to a goal region $x_\text{{goal}} \subseteq X_{\text{free}}$, such that the system respects state constraints at all time. %with respect to system constraints.

\subsubsection{Quadrotor Modeling}
We assume an $SE(3)$ rigid body system of mass $m$ and an inertia matrix $\bm{J}=\bm{J}^{\intercal}\in\mathbb{R}^{3\times3}$ referred to the center of mass (COM) of the quadrotor. Given the state $\bm{x} \in \mathbb{R}^{9}$, the nominal model describing quadrotor's kinematics and dynamics can be expressed as:
\begin{equation}
    \bm{\dot{x}} = f(\bm{x}, \bm{u}) = \begin{bmatrix} \bm{\dot{s}} \\ \bm{\dot{v}} \\ \bm{\dot{\omega}} \end{bmatrix} = \begin{bmatrix} \bm{v} \\ \frac{1}{m} (\bm{q} \odot \bm{f_e}) - \bm{g} \\ \bm{J}^{-1} (\bm{\tau} - \bm{\omega} \times \bm{J}\bm{\omega}) \end{bmatrix} \label{eq:model}
\end{equation}
where $\bm{s}=[x, y, z]^{\intercal}$ is the position vector of the quadrotor COM with respect to the inertial frame origin, $\bm{v}=[v_x, v_y, v_z]^{\intercal}$ is the velocity of the quadrotor COM, $\bm{\omega}=[p, q, r]^{\intercal}$ is the vector of the angular velocity components, $\bm{g}=[0, 0, 9.81]^{\intercal}$ ms$^{-2}$ is the gravitational acceleration, and the quaternion $\bm{q} \in SO(3)$ represents the rotation of the body frame in relation to the world. The total force applied to the quadrotor is $\bm{f_e}$, and the total torque component is $\bm{\tau}$. The control input $\bm{u} \in \mathbb{R}^{4}$ consists of the thrusts delivered by each propeller. It is also noted that $\odot$ represents the quaternion product referring to the vector's rotation by the quaternion.

% \subsection{Problem Statement}
%\subsection{Defining Safety}
\subsection{Safety Definition}
%Converting the system in~\eqref{eq:model} to discrete-time, we get the following formulation:
Consider a system in the following form:
\begin{equation}
    % x_{k+1} = f(x_k, u_k) \label{eq:system}
    %x(k+1) = f(x(k), u(k)) \label{eq:system}
    \dot{x}(t) = f(x(t), u(t)) + d(t) \label{eq:system}
\end{equation}
with state $x(t) \in\mathbb{R}^n$, control input $u(t) \in\mathbb{R}^m$, nominal model $f$, and bounded disturbance $d(t) \in\mathcal{D} \subset\mathbb{R}^n$.
% and $t\in\mathbb{R}_{\geq0}$.
 
The notion of safety for the system in~\eqref{eq:system} is formalized by specifying a safe set $\mathcal{S}\in\mathbb{R}^n$ in which the state of the system must remain at all times to be considered safe during the forward propagation of the system, i.e., $x(t)\in \mathcal{S}$ for all $t\geq0$. A well-established notion studying whether the state of a system belongs to a prescribed set for all time is that of \textit{set invariance} (\cite{blanchini2008set}). Formally, a set $\mathcal{S}\in\mathbb{R}^n$ is said to be \textit{forward invariant} for a closed-loop system with a given control policy $\pi(\cdot)$, i.e.:
\begin{equation}
    \dot{x}(t) = f(x(t), \pi(x(t)) + d(t) \quad t\in\mathbb{R}_{\geq0} \label{Eq2}
\end{equation}
if for any initial condition $x(t_0) \in \mathcal{S}$ we have that $x(t)\in \mathcal{S}$, $\forall t \geq t_0$, where $\mathcal{S} \subseteq \mathcal{C}$, and $\mathcal{C}\in\mathbb{R}^n$ is the set of states that satisfies all the imposed safety constraints, referred to as the constraint set. In this case, the state $x(t_0)$ and the set $\mathcal{S}$ are said to be safe. Furthermore, control laws are said to be safe when they render some nonempty subset of $\mathcal{C}$ forward invariant. Forward invariance, and by extension, safety, are properties of the closed-loop system and are not defined in the absence of a controller. The concept of \textit{control invariance} addresses the question of whether it is possible to find a control policy that will render a particular set forward invariant. Formally, a set $\mathcal{S} \subseteq \mathcal{C}$ is said to be \textit{control invariant} if for any $x(t_0)\in \mathcal{S}$ there exists an input $u(\cdot)$ under the control policy $\pi(\cdot)$ such that $x(t)\in \mathcal{S}$, $\forall t\geq t_0$. This definition implies that a system remains safe under a particular controller only if all solutions to the closed-loop system in~\eqref{Eq2} that begin in the safe set remain in the safe set.

\subsubsection{Robust Invariant Terminal Set (\cite{wabersich2023})}
There exists a control law $\kappa_f : \mathcal{S}_f \rightarrow \mathcal{U}$, and a corresponding robust positively invariant set $\mathcal{S}_f \subseteq \mathcal{X}$, such that for all $x \in \mathcal{S}_f$, it holds that $\kappa_{f}(x) \in \mathcal{U}$ and $f(x, \kappa_{f}(x)) + d(k) \in \mathcal{S}_f$, $\forall d(k) \in\mathcal{D}$, $\forall t\geq t_0$.

\begin{figure*}[htbp]
    \centering
    \begin{subfigure}{0.32\textwidth}
        \includegraphics[width=\linewidth, trim={0 0 0.3cm 1.4cm}]{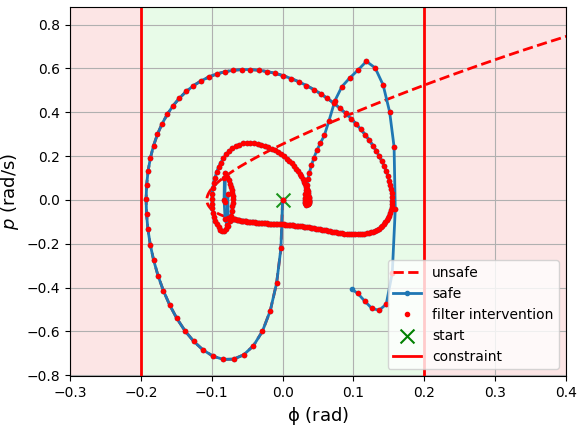}
        \caption{Roll angle constraints and safe set.}
        \label{fig:roll}
    \end{subfigure}
    \hfill
    \begin{subfigure}{0.32\textwidth}
        \includegraphics[width=\linewidth, trim={0 0 0.6cm 1.4cm}]{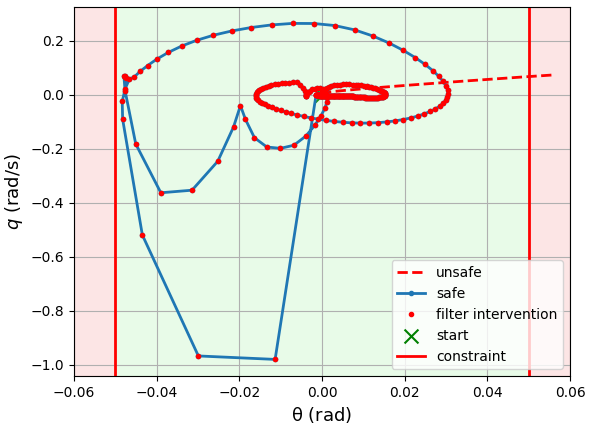}
        \caption{Pitch angle constraints and safe set.}
        \label{fig:pitch}
    \end{subfigure}
    \hfill
    \begin{subfigure}{0.32\textwidth}
        \includegraphics[width=\linewidth]{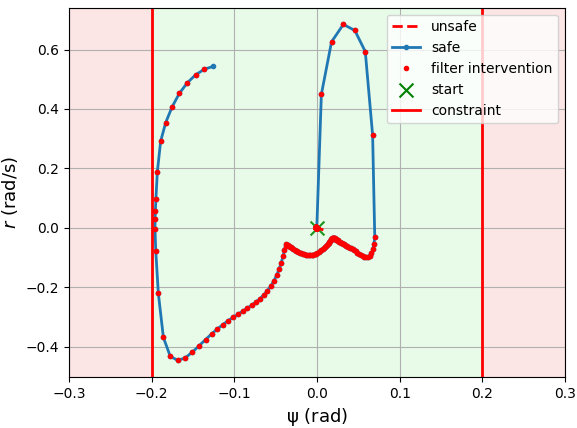}
        \caption{Yaw angle constraints and safe set.}
        \label{fig:yaw}
    \end{subfigure}
    \vspace{0.5em}
    \caption{Simulation results. \textit{Axis} (a) roll, (b) pitch, and (c) yaw. \textit{Shaded green} Safe states, \textit{Shaded red} States for which safety constraints have been violated, \textit{Red line} Boundary of the safe set, \textit{Dashed red} Unsafe states leading to collision, \textit{Dotted blue} States when $u_{\text{des}}(k)$ from the LQR controller is applied, \textit{Dotted red} states when the filtered control $u_{\text{safe}}(k)$ is applied.}
     %\caption{The safe operational region (safe set) in terms of constraints in (a) roll, (b) pitch, and (c) yaw angles of the quadrotor. \textit{Shaded green} safe states, \textit{Shaded red} collision states for which safety constraints have been violated, \textit{Red line} boundary of the safe set, \textit{Dotted blue} states when $u_{\text{des}}(k)$ from the LQR controller is applied, \textit{Dotted red} states when the filtered control $u_{\text{safe}}(k)$ is applied.}
    \label{fig:angles}
\end{figure*}

\subsection{Obtaining Safe Control Inputs} \label{Sec:opti}
For trajectory following, we compute a safe control input sequence and corresponding state trajectory from the one-step ahead predicted state by solving an optimal control problem in a receding horizon fashion. Specifically, consider the system state $x(t_0)$ and the desired control input $u_{\text{des}}(t_0)$ at time $t_0 \in\mathbb{R}_{\geq0}$. The safety of the desired control action $u_{\text{des}}(t)$ is verified by calculating a control input $u(\cdot) \in \mathcal{U}$ over a certain user-specified prediction horizon $T\in\mathbb{R}_{\geq0}$, such that the state of the system at $t_0+T$ lies inside a terminal safe set, i.e., for $t \in [t_0, t_0 + T]$ subject to satisfaction of the system dynamics in~\eqref{eq:system} and $u(t_0) = u_{\text{des}}(t_0), x(t_0 + T) \in \mathcal{S}_T$, where $\mathcal{S}_T$ denotes a terminal safe set at time $t_0 + T$. This strategy is implemented with a receding horizon, implying that the computed safe control input is applied only for the first step, after which a new solution to the optimal control problem will be computed.

Implementing the framework in discrete-time, i.e., given the discrete-time zero-order hold representation of~\eqref{eq:system}:
\begin{equation}
     x(k+1) = f_{\text{d}}(x(k), u(k)) + d(k)
\end{equation}
with $k$ discrete-time index, the safe state trajectory and safe control input are obtained by solving the receding horizon optimization problem:
\begin{subequations}
    \begin{align}
        & \min_{u(i|k)} \| {u_{\text{des}}(k) - u(0|k)} \| \\
        \text{subject to} \quad & x(0|k) = x(k) \\
        & x(T|k) \in \mathcal{S}_T \\
        \text{for} \quad & i = 0, \dots, T-1: \nonumber \\
        & x(i+1|k) = f_{\text{d}}(x(i|k), u(i|k)) + d(k) \\
        & x(i|k) \in \mathcal{C} \quad & \\
        & u(i|k) \in \mathcal{U}
    \end{align}
    \label{eq:optimization}
\end{subequations}
\noindent
\ignorespacesafterend with the goal of minimizing the deviation between the first predicted control input $u(0|k)$ and the currently desired control input $u_{\text{des}}(k)$. Only the first element of the optimal sequence is applied at time $k$, hence $u_{\text{safe}}(k) = \kappa_f(x(k), u_{\text{des}}(k)) = u^{*}(0|k)$, where $\kappa_f(x(k), u_{\text{des}}(k))$ is the proposed safety assurance filter.

\begin{figure*}[htbp]
    \centering
    \begin{subfigure}{0.32\textwidth}
        \includegraphics[width=\linewidth, trim={0 0 1.7cm 1.4cm}]{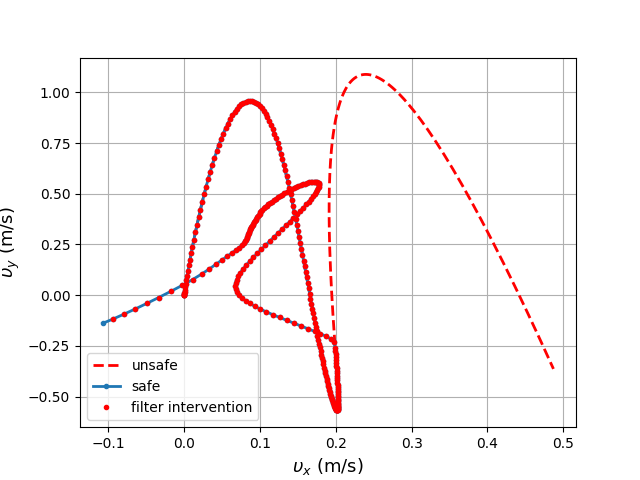}
        \caption{Velocity trajectories in the XY plane.}
        \label{fig:plot_a}
    \end{subfigure}
    \hfill
    \begin{subfigure}{0.32\textwidth}
        \includegraphics[width=\linewidth, trim={0 0 1.7cm 1.4cm}]{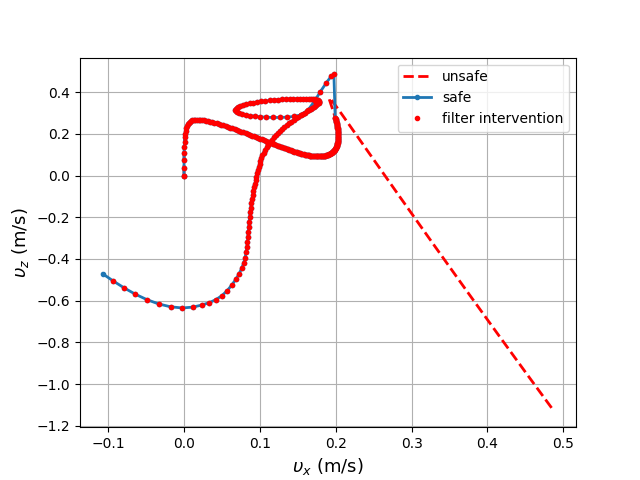}
        \caption{Velocity trajectories in the XZ plane.}
        \label{fig:plot_b}
    \end{subfigure}
    \hfill
    \begin{subfigure}{0.32\textwidth}
        \includegraphics[width=\linewidth]{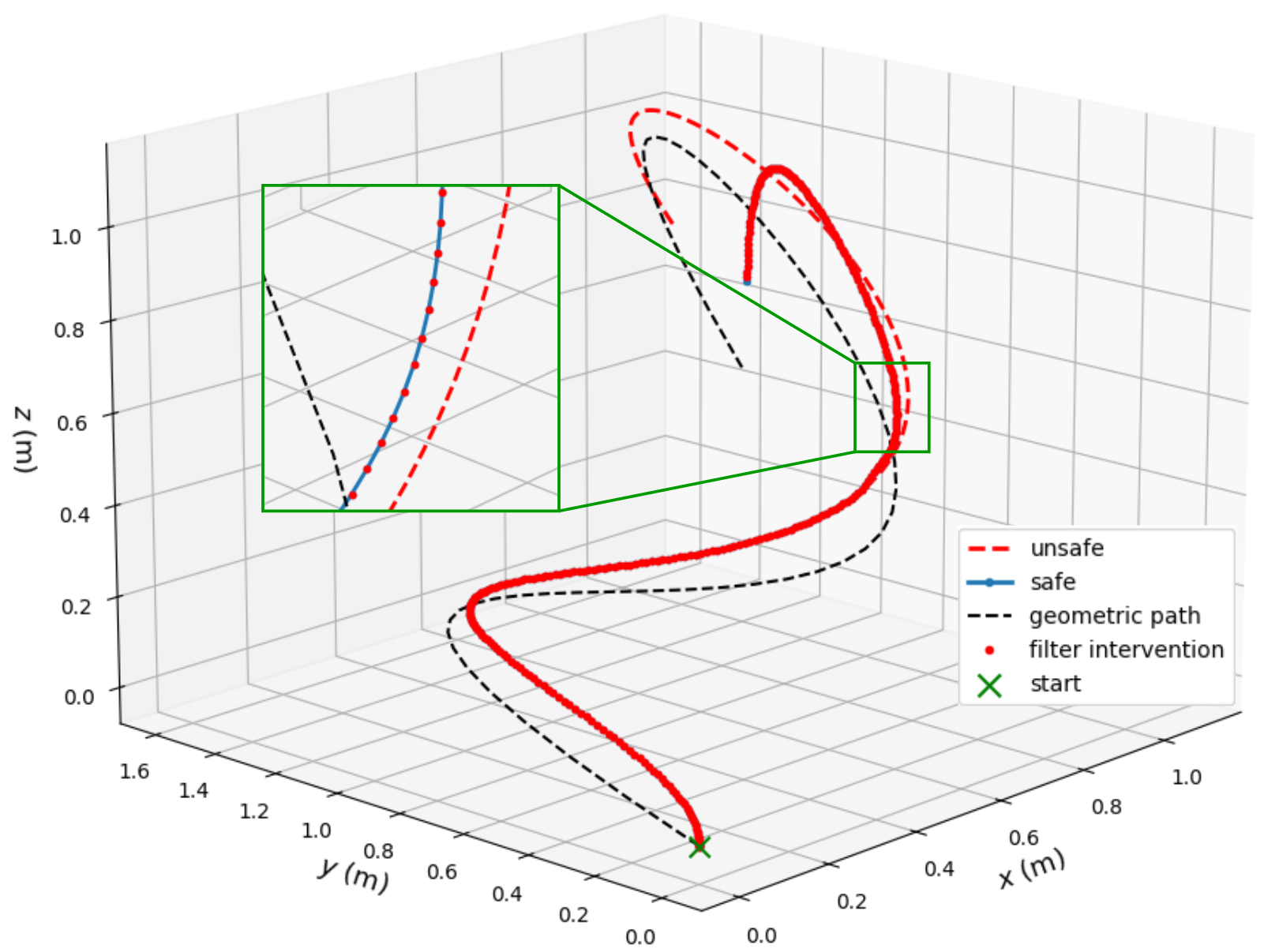}
        \caption{3D position trajectories.}
        \label{fig:3D}
    \end{subfigure}
    \vspace{0.5em}
    \caption{Velocity and position trajectories obtained from the simulation experiments. \textit{Dashed black} Path generated by RRT\textsuperscript{*}, \textit{Dashed red} Unsafe trajectory from LQR leading to a future collision, \textit{Blue line} Final trajectory after safety adjustments, \textit{Dotted red} Interventions by the safety filter to the LQR to adjust for safety constraints.}
    \label{fig:posvel}
\end{figure*}

\begin{figure}[htbp]
\centerline{\includegraphics[width=\linewidth]{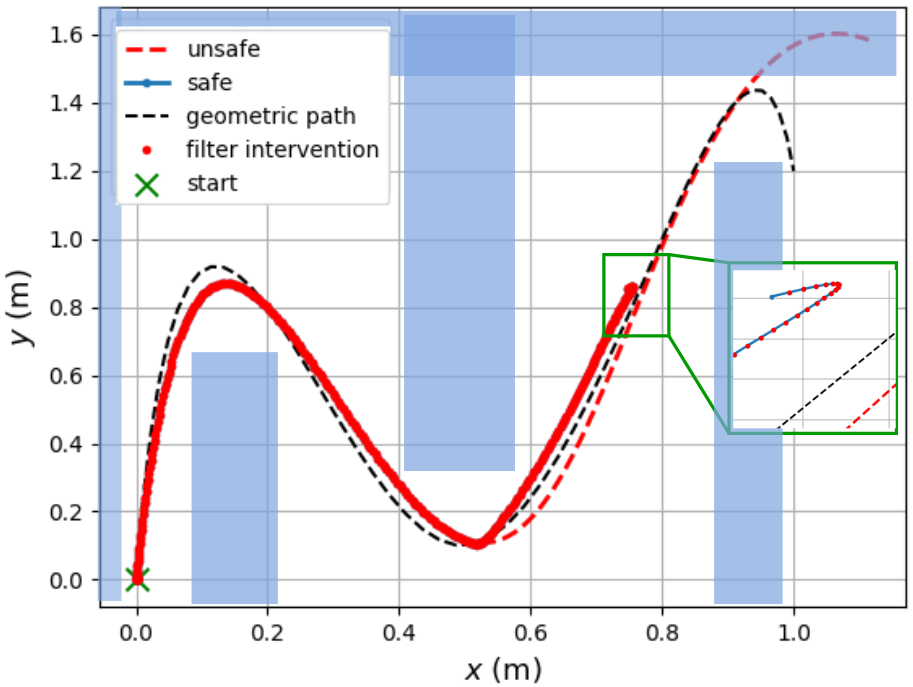}}
\caption{Position trajectories (top view).}
\vspace{0.5em}
\label{fig:plot_c}
\end{figure}

\subsection{Safe Kinodynamic Planning Scheme}
Now we combine the aforementioned safety assurance filter design with a geometric RRT\textsuperscript{*} motion planner. RRT\textsuperscript{*} is an asymptotically optimal planner, which means that given long enough, with probability 1 an optimal solution is found if one exists (\cite{karaman2011sampling}). 
% Unlike RRT, which only looks for a feasible path, RRT\textsuperscript{*} constantly improves the path towards an optimal solution. The final solution path is geometric, meaning that the planner is unaware of the full system dynamics.

\begin{algorithm}[htbp]
\caption{Planning with Safety Assurance}
\label{algo}
\begin{algorithmic}[1]
\Require{$\mathcal{W}, x_{\text{start}}, x_{\text{goal}}, T$}
\Ensure{$\mathcal{P}\textsuperscript{*}$ - trajectory with safety assurance}
\State $\mathcal{P}\textsuperscript{*} \gets \emptyset$
% \While{Goal is not reached}
\ForAll{$j \in (x_{\text{start}}, x_{\text{goal}})$}
    \State $\mathcal{P}(j) \gets \text{RRT\textsuperscript{*}}(\mathcal{W}, x(j), x(j+T))$
    \If{$\mathcal{P}(j)$ is valid}
        \ForAll{$k \in (0, T)$}
            \State Solve \eqref{eq:optimization} to obtain $u_{\text{safe}}(k)$
            \State Apply $u_{\text{safe}}(k)$ to the system
        \EndFor
    \State Obtain $\mathcal{P}_{\text{safe}}(j)$ as safe
    \EndIf
    \State $\mathcal{P}\textsuperscript{*} \gets \mathcal{P}_{\text{safe}}(j)$
    \State $j \gets j + T + 1$
\EndFor
% \EndWhile
\State \text{Obtain $\mathcal{P}\textsuperscript{*}$}
\State \textbf{end} \text{Maintain safety at all times}
\end{algorithmic}
\end{algorithm}

Based on the previously mentioned process, the planner is given a workspace $\mathcal{W}$, an initial position $x_{\text{start}}$, a goal region $x_{\text{goal}}$, and a time horizon $T$. The implementation of the motion planning scheme with safety assurance is presented as pseudocode in Algorithm~\ref{algo}. The output of the algorithm is a trajectory $\mathcal{P}\textsuperscript{*}$ that has been produced with safety assurance. Starting from the initial point of the mission $x_{\text{start}}$, we repeat the safe assurance process by planning ahead for a time period of length $T$ (Lines 2-3). In particular, for every time window $T$, we first request a collision-free path from RRT\textsuperscript{*}, which is first converted to a smooth time-varying trajectory $\mathcal{P}(j)$ (Line 3). If a valid path was returned from the planner, we switch to the safety layer mechanism, where a primary controller (in our case an LQR) and the safety assurance provide a control input $u_{\text{safe}}$ by solving the optimization problem in~\eqref{eq:optimization} (Lines 6-7). We repeat this for every time step $k$ (as presented in Sec.~\ref{Sec:opti}) within the size ($T$) of the respective trajectory $\mathcal{P}(j)$ (Line 5). Therefore, a safe control input $u_{\text{safe}}(k)$ is produced and then applied at every time step $k$. After $u_{\text{safe}}(k)$ has been applied for every time step within $\mathcal{P}(j)$ and the trajectory has been successfully adjusted for safety, we add the corresponding safe trajectory $\mathcal{P}_{\text{safe}}(j)$ to the overall solution trajectory $\mathcal{P}\textsuperscript{*}$ (Line 9). Then, we move on to the next time window of length $T$ requesting a path and following the same procedure until reaching the goal (Line 10). Finally, after all corresponding trajectories have been assured for safety, we obtain $\mathcal{P}\textsuperscript{*}$. A high-level illustration of the algorithm's workflow during one iteration is shown in Fig.~\ref{fig:method}.

%%%%%%%%%%%%%%%%%%%%%%%%%%%%%%%%%%%%%%%%%%%%%%%%%%%%%%%%%%%%%%%%%%%%%%%%%%%%%%%%
\section{SIMULATION RESULTS} \label{sec:results}
To evaluate the performance of our proposed safe kinodynamic planning framework, we conducted experiments in simulation using \texttt{safe-control-gym} (\cite{yuan2022safe}), which provides a physics-based environment using the Crazyflie 2.0 quadrotor model. For the geometric RRT\textsuperscript{*} planner, we use the implementation provided in the Open Motion Planning Library (\texttt{ompl}) (\cite{sucan2012}) for 3D rigid body path planning. The design and optimization part of the safety filter is done using IPOPT (\cite{wachter2006implementation}) and CasADi (\cite{andersson2019casadi}). All the experiments were run on a machine with Intel i7-12700 and 16 GB memory.

We tested the framework in a 3D maze-like environment as shown in Fig.~\ref{fig:env}. The safety filter was evaluated using the previously mentioned quadrotor dynamics over a receding horizon $T = 20$, and controller frequency 50 Hz. The start position was $x_{\text{start}}=[0,0,0]$ m, and the anticipated goal region was $x_{\text{goal}}=[1.0,1.2,0.5]$ m. The workspace $\mathcal{W}$ along with the locations of the obstacles were also given. System constraints were enforced primarily on the roll, pitch, and yaw angles of the quadrotor, specifically, $\phi \in [-0.2, 0.2]$ rad, $\theta \in [-0.05, 0.05]$ rad, and $\psi \in [-0.2, 0.2]$ rad. The translation dynamics (position and velocity) can be indirectly controlled with the roll, pitch, and yaw angles due to the fact that quadrotors are underactuated platforms. Therefore, the presence of constraints in the angles implies that the velocity and position are also constrained.

We present the safe sets in terms of angles and angular velocities in Fig.~\ref{fig:angles}. The safety filter successfully maintains the corresponding parameters within the safe operational region by modifying the control inputs of the LQR to avoid potential unsafe behavior (dashed red trajectory in Figs.~\ref{fig:roll} and \ref{fig:pitch}). Some chattering can be observed due to the repeating adjustments by the safety filter. Similarly, the (adjusted) trajectories of the linear velocity vectors along the XY and XZ planes, and 3D position trajectories are presented in Figs.~\ref{fig:plot_a}, ~\ref{fig:plot_b}, and~\ref{fig:3D} respectively. These trajectories correspond to the end-to-end solution generated by the RRT\textsuperscript{*}, and the safe trajectory $\mathcal{P}\textsuperscript{*}$ constructed using Algorithm~\ref{algo}. Our approach is able to avoid the constraint violation states (dashed red line). Specifically, at some point during the planning process with the quadrotor being at position $[0.5,0.1,0.6]$ m (Fig.~\ref{fig:plot_c}), an unsafe control input is detected by the safety filter. Therefore, the motion plan is adjusted with respect to roll, pitch, and yaw interventions (Figs.~\ref{fig:roll} and~\ref{fig:pitch}), hence, guiding the quadrotor to a slightly modified yet safe goal region. This is better depicted in Fig.~\ref{fig:plot_c} where a top view of the trajectories and environment is shown.

%%%%%%%%%%%%%%%%%%%%%%%%%%%%%%%%%%%%%%%%%%%%%%%%%%%%%%%%%%%%%%%%%%%%%%%%%%%%%%%%
\section{SUMMARY AND CONCLUSION} \label{sec:conclusion}
This work presents a method to maintain safety in kinodynamic planning under system constraints by combining concepts of path planning, optimal control, and run time assurance. We demonstrated its effectiveness in a 3D simulation environment with a quadrotor model by using a safety filter that identifies unsafe control inputs and makes interventions to adjust the planning process based on the given constraints. This process significantly improves the planning scheme in terms of safety robustness. 

Our approach paves the way for interesting follow-up research opportunities and future advancements. We plan to test our framework on a real-world drone testbed. Finally, we plan to investigate the presence of moving obstacles into the system. This would enable the system to quickly adapt to dynamic changes in the environment, thus, making it more robust to real-world planning conditions.

\bibliography{references}
\end{document}